\documentclass[5p,times]{elsarticle}
\usepackage[T1]{fontenc}
\usepackage{amsmath,amssymb}
\usepackage{graphicx}
\usepackage{booktabs}
\usepackage{multirow}
\usepackage{algorithm}
\usepackage{algpseudocode}
\usepackage{xcolor}
\usepackage{microtype}
\usepackage{adjustbox}
\usepackage{xurl}
\usepackage[hidelinks]{hyperref}
\usepackage{siunitx}
\newcommand{\pmsd}[2]{#1\,$\pm$\,#2}
\journal{Future Generation Computer Systems}

\begin{document}
\begin{frontmatter}

\title{Serverless gossip training of LSTM failure detectors: A matched-protocol comparison with federated, local and centralized learning on NASA C-MAPSS}

\author[a]{Yusuf Öztürk}
\author[b]{Enes Göktekin}
\author[a,b]{Bengisu Atlı}
\author[c]{Akın Öztürk}
\author[d]{Zhixiang Wang\corref{cor1}}
\ead{zhixiang.wang@northwestern.edu}
\author[d]{Ulas Bagci\corref{cor1}}
\ead{ulas.bagci@northwestern.edu}
\cortext[cor1]{Co-corresponding authors.}

\affiliation[a]{organization={Department of Electrical and Electronics Engineering, Antalya Bilim University},city={Antalya},country={Türkiye}}
\affiliation[b]{organization={Department of Computer Engineering, Antalya Bilim University},city={Antalya},country={Türkiye}}
\affiliation[c]{organization={Graduate School of Natural and Applied Sciences, Ankara University},city={Ankara},country={Türkiye}}
\affiliation[d]{organization={Department of Radiology, Feinberg School of Medicine, Northwestern University},city={Chicago},state={IL},postcode={60611},country={USA}}

\begin{abstract}
Industrial predictive maintenance increasingly depends on learning from equipment spread across sites whose sensor data cannot easily be pooled. Federated averaging (FedAvg) solves this with a central aggregation server; gossip learning removes the server, but its behaviour for recurrent failure-detection models has not been measured under controlled conditions. We compare synchronous ring gossip with FedAvg, isolated local training and a centralized reference for a stacked LSTM that detects imminent failure on the NASA C-MAPSS turbofan benchmark. All methods share one open implementation, architecture, initialization, optimizer, data split and training budget, and the primary endpoint uses one terminal window per test engine to avoid the statistical dependence of overlapping windows. On FD001 (five seeds), gossip reached a terminal-window F1 of 89.6\,$\pm$\,1.3\%, compared with 89.9\,$\pm$\,1.1\% for FedAvg, 83.6\,$\pm$\,6.7\% for local training and 93.5\,$\pm$\,2.1\% for centralized training, while transmitting the same payload as FedAvg without a coordinator. Node models agreed closely but not exactly (1.8\% pairwise decision disagreement versus 5.6\% without communication). Across FD002--FD004, peer communication improved terminal-window F1 over local training by 13--28 points; gossip matched FedAvg on FD003 and FD004 but was 4.3 points lower on the multi-condition FD002 subset. Simulated message loss, node failure and server outage changed neither method appreciably, whereas larger rings degraded gossip faster. Ring gossip is therefore a practical serverless alternative when data heterogeneity is moderate, and faster-mixing topologies become important as heterogeneity grows.
\end{abstract}


\end{frontmatter}

\section{Introduction}
\label{sec:intro}

Predictive maintenance (PdM) uses condition-monitoring data to anticipate failures and schedule interventions before breakdowns occur, reducing unplanned downtime and maintenance cost \cite{lee2014phm,jardine2006review}. Deep sequence models have become standard tools for this task because they learn degradation patterns directly from multivariate sensor streams \cite{wang2018dlsmart,zheng2017lstm,li2018cnn,wu2024survey}. Their accuracy, however, depends on the amount and diversity of run-to-failure data, which in practice is distributed over plants, fleets or operators that are often unwilling or unable to centralize it \cite{tao2018datadriven,nunes2023challenges,mallioris2024pdm}.

Federated learning (FL) addresses this by training a shared model while raw data remain on the participating devices \cite{mcmahan2017fedavg,kairouz2021advances}. It has been applied to fault diagnosis \cite{mehta2023fl}, to anomaly detection under distribution shift \cite{ahn2023fl}, and to remaining-useful-life (RUL) prognostics across airlines \cite{landau2026fl}, and a federated benchmark on C-MAPSS was recently released \cite{sorrenti2026fedcmapss}. All of these systems rely on a server that collects and redistributes models in every round. In edge and industrial settings this coordinator is a single point of failure, a communication hub whose load grows with the number of participants, and an organizational obstacle when no party is trusted to host it. Gossip learning removes the server: each node averages its model with a few neighbours, and information spreads through the network over successive rounds \cite{kempe2003gossip,boyd2006gossip,ormandi2013gossip}. Decentralized stochastic gradient descent can, under suitable conditions, match the convergence of its centralized counterpart \cite{lian2017dpsgd,koloskova2020unified}, and gossip learning has been shown to be competitive with FL on several benchmarks \cite{hegedus2021gossip}.

For practitioners designing a serverless maintenance system, three questions remain open. First, does peer-to-peer communication actually improve on what each site could learn alone, and by how much? Second, how much accuracy does removing the server cost relative to FedAvg when both are implemented and trained identically? Third, how do the answers change under faults, heterogeneous data and larger networks? Answering them requires a controlled comparison: many PdM studies compare methods implemented in different code bases with different budgets, and many evaluate on every sliding window of a small number of test engines, which treats strongly overlapping and therefore dependent windows as independent observations.

This paper provides such a comparison for LSTM-based imminent-failure detection on the NASA C-MAPSS turbofan benchmark \cite{saxena2008cmapss}. Our contributions are as follows:
\begin{itemize}
\item A matched-protocol comparison of synchronous ring gossip, FedAvg, isolated local training and centralized training, in which all methods share one implementation, architecture, initialization, optimizer, data split and training budget, evaluated over repeated seeds with paired tests.
\item An evaluation design that uses one terminal window per test engine as the primary endpoint and resamples whole engines for all confidence intervals, avoiding inflated precision from overlapping windows.
\item Direct measurements of how closely gossip node models agree, together with an exact communication ledger that includes an event-triggered gossip variant.
\item Simulations of message loss, node failure, server outage, non-IID partitioning and network size, and a replication on all four C-MAPSS subsets, which identify where ring gossip matches FedAvg and where its slower information mixing becomes costly.
\item Open code, data and per-run results from which every reported number can be regenerated.
\end{itemize}

Section~\ref{sec:related} reviews related work. Section~\ref{sec:methods} describes the learning task, the training protocols and the communication model, and Section~\ref{sec:setup} the data and evaluation design. Section~\ref{sec:results} reports the results, Section~\ref{sec:discussion} discusses their implications and limitations, and Section~\ref{sec:conclusion} concludes.

\section{Related work}
\label{sec:related}

\subsection{Deep learning for failure prognostics}
Data-driven prognostics estimate either the remaining useful life of an asset or the probability that it will fail within a maintenance horizon \cite{si2011rul,heng2009rotating,carvalho2019slr}. On C-MAPSS, LSTM networks \cite{hochreiter1997lstm,gers2000forget,zheng2017lstm} and convolutional networks \cite{li2018cnn} are established baselines, and recent surveys summarize the wide range of deep architectures that have since been proposed \cite{wu2024survey}. Edge computing platforms increasingly perform part of this processing close to the monitored equipment \cite{mourtzis2022edge}. Our focus is not a new predictor: we deliberately use a conventional two-layer LSTM so that differences between conditions can be attributed to how models are trained and combined across sites.

\subsection{Federated learning for maintenance}
FedAvg alternates local training on each client with server-side averaging of model parameters \cite{mcmahan2017fedavg}; its extensions address statistical heterogeneity, communication efficiency and privacy \cite{kairouz2021advances,rauniyar2024fl}. In manufacturing and prognostics, FL has been used for mixed fault diagnosis in rotating machinery \cite{mehta2023fl}, predictive maintenance and anomaly detection under data-distribution shifts \cite{ahn2023fl}, and collaborative RUL prognostics among airlines on N-CMAPSS with robust aggregation and decentralized validation \cite{landau2026fl}. The FedCMAPSS benchmark standardizes federated RUL tasks on C-MAPSS from IID to strongly heterogeneous client settings \cite{sorrenti2026fedcmapss}. These works establish that collaborative training is valuable for prognostics, but they all assume a central aggregator.

\subsection{Gossip and decentralized learning}
Gossip protocols compute network-wide aggregates through repeated local exchanges \cite{kempe2003gossip,boyd2006gossip}. For averaging with a fixed doubly stochastic mixing matrix, disagreement contracts at a rate governed by the second-largest eigenvalue modulus (SLEM) of that matrix \cite{xiao2004averaging,boyd2006gossip}, and distributed subgradient methods combine such averaging with local optimization steps \cite{nedic2009subgradient}. Decentralized parallel SGD can match centralized SGD when the network mixes well enough \cite{lian2017dpsgd}, and a unified analysis covers local updates and changing topologies \cite{koloskova2020unified}. Gossip learning applies these ideas to machine-learning models without any coordinator \cite{ormandi2013gossip}, and a large empirical study found it competitive with FL across several tasks \cite{hegedus2021gossip}. Theory and general benchmarks therefore suggest that a well-connected gossip network can approach FedAvg; how a sparse ring behaves for recurrent failure detectors, under realistic data fragmentation and faults, is what the present study measures.

\section{Methods}
\label{sec:methods}

\subsection{Learning task}
\label{sec:task}
Each engine produces a multivariate time series of operational settings and sensor readings, one vector per operating cycle. For an engine that fails at cycle $T$, the remaining useful life at cycle $t$ is $\mathrm{RUL}(t)=T-t$. An input window $\mathbf{X}_t\in\mathbb{R}^{50\times 25}$ contains the 50 consecutive cycles ending at $t$, and its label is
\begin{equation}
y_t=\mathbb{1}\left[\mathrm{RUL}(t)\le H\right],\qquad H=30\ \text{cycles},
\label{eq:label}
\end{equation}
so that a positive prediction is an alarm that failure is expected within the maintenance horizon $H$. The horizon follows common practice for this benchmark and was fixed before any experiment. A model $f_\theta$ outputs the probability $\hat{y}_t=f_\theta(\mathbf{X}_t)$ and is trained with the binary cross-entropy loss; an alarm is raised when $\hat{y}_t\ge 0.5$.

The classifier is a two-layer LSTM \cite{hochreiter1997lstm,gers2000forget} with 100 and 50 hidden units, dropout 0.2 after each layer, and a sigmoid output unit, giving 80,651 trainable parameters. The first layer returns its full hidden-state sequence and the second only its final state. The cell equations are given in Supplementary Section~S1.

\subsection{Training protocols}
\label{sec:protocols}
\begin{figure*}[t]
\centering
\includegraphics[width=\textwidth]{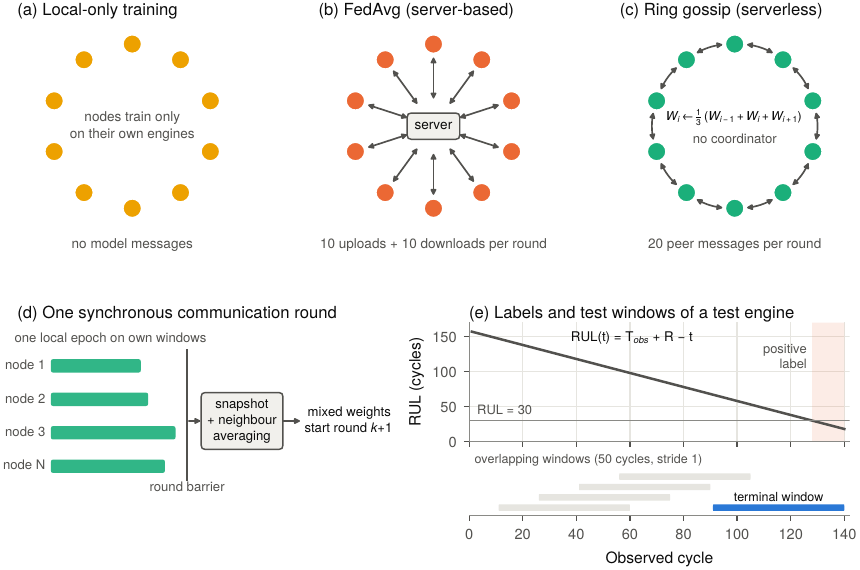}
\caption{Training protocols and evaluation design. (a)--(c) Local-only training, FedAvg and ring gossip on $N=10$ nodes. (d) Each communication round consists of one local epoch per node, a synchronization barrier and one mixing step. (e) For a truncated test engine with official terminal RUL $R$ observed up to cycle $T_{\mathrm{obs}}$, labels follow from $\mathrm{RUL}(t)=T_{\mathrm{obs}}+R-t$; the primary endpoint scores only the terminal window of each engine, whereas the secondary analysis scores all overlapping windows.}
\label{fig:overview}
\end{figure*}

Training data are distributed across $N$ nodes (edge sites), each holding the complete histories of a disjoint set of training engines. All protocols start from the same seed-specific initial parameters, copied to every node, and proceed for $K=50$ rounds; in each round every node performs one epoch of local training on its own windows (Fig.~\ref{fig:overview}). The protocols differ only in what happens after local training.

\paragraph{Local-only} Nodes never communicate. This condition measures what each site can learn on its own and therefore quantifies the value of collaboration.

\paragraph{FedAvg} After local training, nodes upload their parameters to a server, which returns the average weighted by the number of training windows at each node \cite{mcmahan2017fedavg}.

\paragraph{Ring gossip} Nodes are arranged in a ring, and node $i$ exchanges parameters only with nodes $i-1$ and $i+1$ (indices modulo $N$). After local training produces $\tilde{W}_i^{(k)}$, all nodes take a snapshot and mix synchronously:
\begin{equation}
W_i^{(k+1)}=\tfrac{1}{3}\left(\tilde{W}_{i-1}^{(k)}+\tilde{W}_i^{(k)}+\tilde{W}_{i+1}^{(k)}\right).
\label{eq:gossip}
\end{equation}
The mixed parameters are the starting point of the next round (Algorithm~\ref{alg:gossip}). No coordinator is involved at any stage.

\paragraph{Centralized} A single model is trained on the union of all training windows. It is not deployable when data cannot be pooled and serves as a reference.

\paragraph{Variants} We additionally evaluate (i) class-weighted FedAvg and gossip, which scale the positive-class loss by the training negative-to-positive ratio (capped at 20), and (ii) event-triggered gossip, in which a node initiates an exchange only if the relative $\ell_2$ change of its parameters since its last exchange is at least 2\% or it has been silent for five rounds; an edge is used if either endpoint triggers, and a 16-byte trigger message is counted on every directed link in every round.

\begin{algorithm}[t]
\caption{Synchronous ring gossip training}
\label{alg:gossip}
\begin{algorithmic}[1]
\Require nodes $i=1,\dots,N$ with local windows $\mathcal{D}_i$; common initialization $W^{(0)}$; rounds $K$
\State $W_i^{(0)}\gets W^{(0)}$ for all $i$
\For{$k=0,\dots,K-1$}
  \For{each node $i$ \textbf{in parallel}}
    \State $\tilde{W}_i^{(k)}\gets$ one epoch of Adam on $\mathcal{D}_i$ starting from $W_i^{(k)}$
  \EndFor
  \State wait until all nodes have finished \Comment{round barrier}
  \For{each node $i$}
    \State send $\tilde{W}_i^{(k)}$ to nodes $i-1$ and $i+1$
    \State $W_i^{(k+1)}\gets\frac{1}{3}\big(\tilde{W}_{i-1}^{(k)}+\tilde{W}_i^{(k)}+\tilde{W}_{i+1}^{(k)}\big)$
  \EndFor
\EndFor
\State \Return node models $W_1^{(K)},\dots,W_N^{(K)}$
\end{algorithmic}
\end{algorithm}

\subsection{Mixing and consensus}
\label{sec:mixing}
Stacking the node parameters, Eq.~\eqref{eq:gossip} is $\mathbf{W}^{(k+1)}=A\,\tilde{\mathbf{W}}^{(k)}$ with a symmetric, doubly stochastic circulant matrix $A$ whose rows contain $1/3$ at positions $i-1$, $i$ and $i+1$. Its eigenvalues are $\lambda_j=\frac{1}{3}\left(1+2\cos(2\pi j/N)\right)$, $j=0,\dots,N-1$. For pure averaging without local training, the squared deviation from the network mean contracts by at least $\mathrm{SLEM}(A)^2$ per round \cite{xiao2004averaging,boyd2006gossip}; for $N=10$, $\mathrm{SLEM}(A)=0.873$, and for $N=40$ it rises to 0.992. Because every round also applies local gradient steps on different data, which move the models apart again, this contraction does not imply that trained networks reach consensus \cite{lian2017dpsgd,koloskova2020unified}. We therefore measure agreement directly (Section~\ref{sec:res-agreement}): as the root-mean-square distance of node parameters from their mean, and as the fraction of node pairs whose final models make different decisions on a fixed probe of validation windows (eight per validation engine).

\subsection{Communication accounting}
\label{sec:comm}
Communication is counted per message. A model message carries 80,651 float32 parameters, i.e. 322,604 bytes. With $N=10$, gossip sends $2N=20$ directed messages per round and FedAvg sends $N$ uploads and $N$ downloads, so both transmit $20\times 50\times 322{,}604=322{,}604{,}000$ bytes (307.66\,MiB) over 50 rounds. Total payload is therefore equal by construction; the protocols differ in its distribution, since each gossip node talks to two peers whereas the FedAvg server terminates all $2N$ transfers. Headers, serialization, acknowledgements, initial model distribution and the exchange of normalization statistics are not counted.

\section{Experimental setup}
\label{sec:setup}

\subsection{Data and labels}
\label{sec:data}
C-MAPSS contains simulated run-to-failure trajectories of turbofan engines with three operational settings and 21 sensors per cycle \cite{saxena2008cmapss}. Each of its four subsets provides complete training trajectories, test trajectories truncated at an unknown point before failure, and the true RUL at the last observed test cycle (Table~\ref{tab:data}). FD001 is used for all main analyses; FD002--FD004 add multiple operating conditions and a second fault mode.

\begin{table}[t]
\centering
\caption{C-MAPSS subsets and evaluation populations. Test engines shorter than the 50-cycle window are excluded; positives are windows with RUL $\le$ 30 cycles.}
\label{tab:data}
\footnotesize
\setlength{\tabcolsep}{3pt}
\begin{adjustbox}{max width=\linewidth}
\begin{tabular}{@{}lccccc@{}}
\toprule
Subset & Cond./faults & Train eng. & Test eng. (pos.) & Windows (pos.) & Seeds \\
\midrule
FD001 & 1 / 1 & 100 & 93 (25) & 8,255 (332) & 5 \\
FD002 & 6 / 1 & 260 & 235 (61) & 21,584 (1,087) & 3 \\
FD003 & 1 / 2 & 100 & 97 (20) & 11,717 (291) & 3 \\
FD004 & 6 / 2 & 249 & 228 (53) & 29,416 (864) & 3 \\
\bottomrule
\end{tabular}
\end{adjustbox}
\end{table}

For a test engine observed up to cycle $T_{\mathrm{obs}}$ with official terminal RUL $R$, the failure cycle is $T_{\mathrm{obs}}+R$, so every cycle of its truncated history receives $\mathrm{RUL}(t)=T_{\mathrm{obs}}+R-t$ (Fig.~\ref{fig:overview}e). Labels are aligned to the last cycle of each window and RUL values are never used as inputs. Each time step has 25 features: the three operational settings, the cycle index and the 21 sensors. Within each seed, 80\% of the training engines are used for training and 20\% for validation, split at the engine level. Min--max scaling is fitted on the training engines only and applied unchanged to validation and test data.

Training engines are assigned at random to $N=10$ nodes, so that each node holds eight FD001 training engines. Validation and test engines are assigned to nodes independently at random; in the decentralized protocols each test engine is scored by the final model of its assigned node, without ensembling. Pooled metrics are computed over all nodes' predictions.

\subsection{Evaluation design and statistics}
\label{sec:eval}
Consecutive windows of the same engine share up to 49 of 50 cycles and are strongly dependent. The \emph{primary endpoint} therefore scores exactly one window per test engine, the terminal window (93 windows on FD001). As a secondary analysis, every valid window is scored with stride~1, describing behaviour across early, mid and late degradation. For both populations we report positive-class F1 and average precision (AP) at the fixed threshold of 0.5; accuracy is uninformative because only 4\% of FD001 windows are positive, and precision, recall and accuracy are listed in Supplementary Table~S1.

Results are mean\,$\pm$\,standard deviation over seeds. A seed jointly determines the train/validation split, the node assignment, the initialization and the minibatch order, and is shared across methods, so comparisons are paired. We report paired per-seed differences with exact two-sided sign-flip tests; with five seeds the smallest attainable $p$-value is 0.0625, so we interpret effect sizes rather than significance. Within-seed uncertainty is quantified with 95\% bootstrap intervals that resample whole test engines (500 replicates), never individual windows.

\subsection{Implementation}
\label{sec:impl}
All protocols use Adam \cite{kingma2015adam} with learning rate $10^{-3}$, batch size 200, gradient-norm clipping at 5 and 50 rounds without early stopping; the final models are evaluated. The optimizer state is reset at the start of each round for every protocol, including centralized training, so that all methods restart from their (possibly mixed) parameters in the same way. Input weights use Xavier initialization, recurrent weights orthogonal initialization and forget-gate biases one. Two differences between protocols are unavoidable: centralized training takes fewer, larger-population optimizer steps than the sum of local steps, and FedAvg weights nodes by sample count whereas ring mixing weights neighbours equally. The network is simulated synchronously in a single process on CPU with deterministic algorithms (PyTorch~2). FD001 experiments were run on one machine; FD002--FD004 were run on a separate Linux server (Python 3.9, PyTorch 2.5.1), and methods are compared only within a subset.

\subsection{Additional scenarios}
\label{sec:scenarios}
With three seeds (11, 22, 33) on FD001, FedAvg and ring gossip were further compared under: independent loss of 20\% of directed messages, where a gossip exchange is applied only if both directions arrive; permanent failure of one node from round 25, whose last model continues to score its test engines; a server outage between rounds 20 and 35, during which FedAvg nodes continue training locally; a non-IID partition in which training engines are sorted by lifetime and assigned to nodes in contiguous blocks; removal of the cycle-index input; and $N=5$, 20 and 40 nodes with the total training data held fixed. Finally, centralized, FedAvg, gossip and local-only training were repeated on FD002, FD003 and FD004 with three seeds and no change to the protocol or hyperparameters.

As an exploratory privacy diagnostic, a loss-threshold membership test scores each training engine (member) and validation engine (non-member) by the mean loss of the final model over its last 50 windows; the area under the ROC curve (AUC) measures how well members are distinguished (0.5 corresponds to chance) \cite{shokri2017mia}.

\section{Results}
\label{sec:results}

\subsection{Main comparison on FD001}
\label{sec:res-main}

\begin{figure*}[t]
\centering
\includegraphics[width=\textwidth]{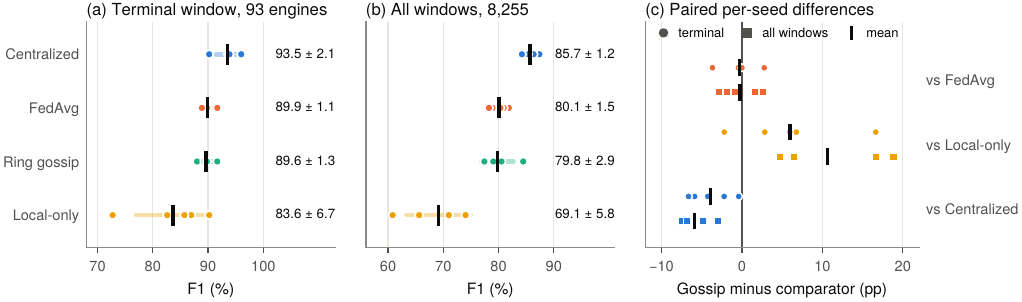}
\caption{Main comparison on FD001. (a) Terminal-window and (b) all-window F1 for each seed (dots), with the mean (black tick) and $\pm$1 SD (shaded bar). (c) Paired per-seed differences between ring gossip and each comparator on the terminal (circles) and all-window (squares) populations.}
\label{fig:main}
\end{figure*}

Table~\ref{tab:main} and Fig.~\ref{fig:main} summarize the matched comparison. On the primary endpoint, centralized training reached 93.5\,$\pm$\,2.1\% F1, FedAvg 89.9\,$\pm$\,1.1\% and ring gossip 89.6\,$\pm$\,1.3\%. The paired gossip$-$FedAvg difference was $-0.3$ percentage points (pp; SD 2.3; $p=0.75$), smaller than the variation between seeds. Local-only training was lower and much less stable (83.6\,$\pm$\,6.7\%), and gossip exceeded it by 6.0\,pp on average. On all windows the ordering was the same: 85.7\,$\pm$\,1.2\% (centralized), 80.1\,$\pm$\,1.5\% (FedAvg), 79.8\,$\pm$\,2.9\% (gossip) and 69.1\,$\pm$\,5.8\% (local-only). Here gossip exceeded local-only training in every seed (+10.7\,pp), and the threshold-free AP showed the same pattern (90.6\% for gossip, 90.9\% for FedAvg, 74.0\% for local-only).

\begin{table*}[t]
\centering
\caption{Main results on FD001 (five seeds, mean\,$\pm$\,SD, threshold 0.5). Terminal: one window per test engine (93 engines, 25 positive). All windows: 8,255 windows (332 positive). Disagreement: fraction of node pairs whose final models give different decisions on the validation probe. Payload: model bytes exchanged over 50 rounds.}
\label{tab:main}
\footnotesize
\begin{adjustbox}{max width=\linewidth}
\begin{tabular}{@{}lcccccc@{}}
\toprule
& \multicolumn{2}{c}{Terminal window (primary)} & \multicolumn{2}{c}{All windows (secondary)} & & \\
\cmidrule(lr){2-3}\cmidrule(lr){4-5}
Method & F1 (\%) & AP (\%) & F1 (\%) & AP (\%) & Disagreement (\%) & Payload (MiB) \\
\midrule
Centralized (reference) & \pmsd{93.5}{2.1} & \pmsd{99.4}{0.5} & \pmsd{85.7}{1.2} & \pmsd{94.9}{1.6} & -- & 0 \\
FedAvg & \pmsd{89.9}{1.1} & \pmsd{98.5}{0.7} & \pmsd{80.1}{1.5} & \pmsd{90.9}{1.2} & 0.0 & 307.7 \\
Ring gossip & \pmsd{89.6}{1.3} & \pmsd{97.7}{0.1} & \pmsd{79.8}{2.9} & \pmsd{90.6}{2.1} & 1.8 & 307.7 \\
Local-only & \pmsd{83.6}{6.7} & \pmsd{91.3}{5.6} & \pmsd{69.1}{5.8} & \pmsd{74.0}{12.7} & 5.6 & 0 \\
\midrule
FedAvg, class-weighted & \pmsd{94.9}{1.7} & \pmsd{99.1}{0.4} & \pmsd{80.0}{1.1} & \pmsd{91.1}{1.7} & 0.0 & 307.7 \\
Gossip, class-weighted & \pmsd{92.1}{2.8} & \pmsd{97.5}{1.3} & \pmsd{77.4}{3.1} & \pmsd{89.4}{2.9} & 2.1 & 307.7 \\
Event-triggered gossip & \pmsd{89.1}{0.7} & \pmsd{97.8}{0.4} & \pmsd{80.0}{3.1} & \pmsd{90.6}{2.3} & 1.9 & 307.1 \\
Event-triggered gossip, class-weighted & \pmsd{91.2}{2.5} & \pmsd{97.5}{1.3} & \pmsd{78.0}{2.7} & \pmsd{89.6}{2.8} & 2.2 & 307.3 \\
\bottomrule
\end{tabular}
\end{adjustbox}
\end{table*}

The test set limits how finely strong methods can be separated. With 25 positive terminal windows, one missed engine changes recall by 4\,pp, and engine-bootstrap intervals within a single seed were correspondingly wide: for seed 11, terminal F1 was 88.0\% (95\% CI 76.0--96.1\%) for gossip and 91.7\% (81.5--98.0\%) for FedAvg, and the mean interval width across seeds was 19.0\,pp for gossip and 19.7\,pp for FedAvg. All-window intervals were narrower (14.7 and 13.8\,pp) but still substantial, which illustrates why treating 8,255 overlapping windows as independent would overstate precision.

Class weighting moved the operating point toward higher recall on terminal windows but did not improve all-window F1, indicating that it mainly traded missed failures for false alarms at the fixed threshold. When the threshold was instead selected on the validation engines to maximize F1, the ordering of the unweighted methods was unchanged (centralized 95.5\,$\pm$\,1.8\%, FedAvg 91.6\,$\pm$\,1.6\%, gossip 89.7\,$\pm$\,2.2\%, local-only 85.7\,$\pm$\,3.7\%) and the advantage of class weighting shrank or disappeared (FedAvg 93.5\,$\pm$\,0.8\%, gossip 89.2\,$\pm$\,3.9\%).

\subsection{Agreement between node models}
\label{sec:res-agreement}
\begin{figure*}[t]
\centering
\includegraphics[width=\textwidth]{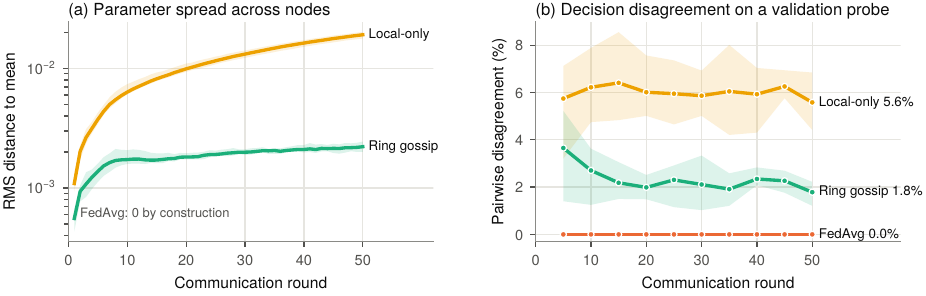}
\caption{Agreement between node models on FD001 (mean over five seeds; shaded band: range across seeds). (a) Root-mean-square distance of node parameters from the network mean (log scale). (b) Pairwise decision disagreement of node models on a fixed probe of validation windows, evaluated every five rounds.}
\label{fig:agreement}
\end{figure*}

Without communication, the parameter spread between nodes grew steadily throughout training, and the final local models disagreed on 5.6\% of probe decisions (Fig.~\ref{fig:agreement}). Ring gossip kept the spread about an order of magnitude smaller and reduced disagreement to 1.8\,$\pm$\,0.5\% after 50 rounds; FedAvg overwrites all node models with the server average and has no disagreement by construction. Gossip therefore produces closely agreeing but not identical models: the residual disagreement reflects local steps taken after the last mixing step and the slow mixing of a ten-node ring ($\mathrm{SLEM}=0.873$). In deployment, the same engine could receive a slightly different risk score depending on which node evaluates it.

\subsection{Communication}
\label{sec:res-comm}
FedAvg and ring gossip each exchanged 1,000 model messages (307.66\,MiB) over 50 rounds, as derived in Section~\ref{sec:comm}. The event-triggered variant transmitted 307.1\,$\pm$\,0.9\,MiB including trigger messages and reached F1 similar to standard gossip. With the change threshold fixed in advance at 2\%, the relative parameter change after one local epoch remained above the threshold in almost every round, so nearly every node triggered every round and no communication was saved. Larger thresholds or several local epochs between exchanges would be needed for savings; we did not tune the threshold on test data. Because all protocols were simulated sequentially on one CPU, wall-clock time reflects the number of local passes rather than deployment speed and is not used as a cost measure.

\subsection{Faults, heterogeneity and network size}
\label{sec:res-scenarios}
\begin{figure*}[t]
\centering
\includegraphics[width=\textwidth]{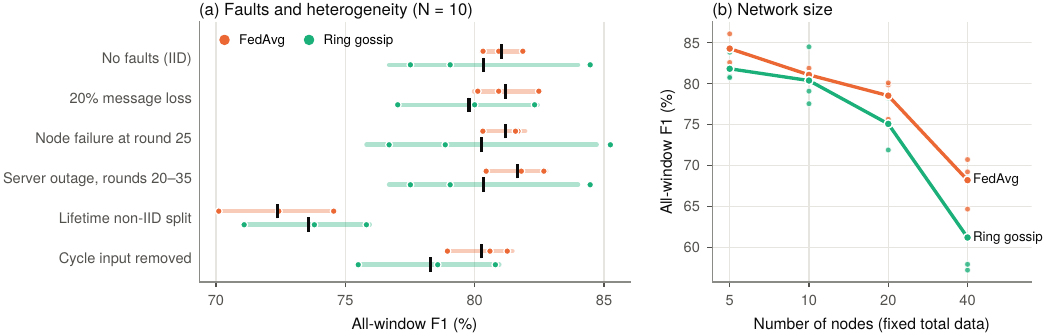}
\caption{FedAvg and ring gossip under additional scenarios on FD001 (three seeds). (a) All-window F1 under faults, a lifetime-sorted non-IID partition and removal of the cycle input with $N=10$ (dots: seeds; black tick: mean; bar: $\pm$1 SD). (b) All-window F1 when the same training data are split across $N=5$ to 40 nodes.}
\label{fig:exploratory}
\end{figure*}

Table~\ref{tab:scenarios} and Fig.~\ref{fig:exploratory} report the additional FD001 scenarios. Neither protocol degraded appreciably under 20\% message loss (all-window F1 81.2\% for FedAvg and 79.8\% for gossip, compared with 81.0\% and 80.3\% without faults), after the permanent failure of one node (81.2\% and 80.3\%), or during a temporary server outage, in which FedAvg nodes continued training locally and re-synchronized afterwards (81.6\%). With the lifetime-sorted non-IID partition, both collaborative protocols lost 7--9\,pp relative to the IID split; gossip (73.6\%) was not worse than FedAvg (72.4\%), and both remained well above local-only training (65.0\%). Removing the cycle-index input reduced F1 slightly (80.3\% and 78.3\%), showing that the models do not rely primarily on elapsed time.

\begin{table*}[t]
\centering
\caption{FedAvg versus ring gossip under additional scenarios on FD001 (three seeds, mean\,$\pm$\,SD, \%). Payload in MiB over 50 rounds.}
\label{tab:scenarios}
\footnotesize
\setlength{\tabcolsep}{6pt}
\begin{adjustbox}{max width=\linewidth}
\begin{tabular}{@{}lcccccc@{}}
\toprule
& \multicolumn{2}{c}{Terminal F1} & \multicolumn{2}{c}{All-window F1} & \multicolumn{2}{c}{Payload} \\
\cmidrule(lr){2-3}\cmidrule(lr){4-5}\cmidrule(lr){6-7}
Scenario & FedAvg & Gossip & FedAvg & Gossip & FedAvg & Gossip \\
\midrule
No faults (IID) & \pmsd{90.3}{1.2} & \pmsd{88.9}{0.8} & \pmsd{81.0}{0.8} & \pmsd{80.3}{3.6} & 307.7 & 307.7 \\
Message loss 20\% & \pmsd{89.5}{2.4} & \pmsd{91.0}{1.1} & \pmsd{81.2}{1.2} & \pmsd{79.8}{2.7} & 307.7 & 307.7 \\
Node failure (r.\,25) & \pmsd{91.0}{3.2} & \pmsd{87.8}{3.8} & \pmsd{81.2}{0.8} & \pmsd{80.3}{4.4} & 292.3 & 276.9 \\
Server outage (r.\,20--35) & \pmsd{92.5}{1.2} & \pmsd{88.9}{0.8} & \pmsd{81.6}{1.1} & \pmsd{80.3}{3.6} & 215.4 & 307.7 \\
Lifetime non-IID & \pmsd{80.0}{1.6} & \pmsd{84.1}{1.3} & \pmsd{72.4}{2.2} & \pmsd{73.6}{2.4} & 307.7 & 307.7 \\
No cycle input & \pmsd{89.5}{0.3} & \pmsd{89.8}{1.8} & \pmsd{80.3}{1.2} & \pmsd{78.3}{2.7} & 307.7 & 307.7 \\
$N=5$ & \pmsd{93.8}{2.2} & \pmsd{90.0}{2.6} & \pmsd{84.3}{1.7} & \pmsd{81.8}{1.8} & 153.8 & 153.8 \\
$N=20$ & \pmsd{89.3}{2.4} & \pmsd{89.2}{0.3} & \pmsd{78.5}{2.5} & \pmsd{75.1}{3.2} & 615.3 & 615.3 \\
$N=40$ & \pmsd{78.3}{4.8} & \pmsd{75.2}{6.1} & \pmsd{68.2}{3.1} & \pmsd{61.2}{6.3} & 1230.6 & 1230.6 \\
\bottomrule
\end{tabular}
\end{adjustbox}
\end{table*}

Splitting the same data across more nodes reduced F1 for both protocols, but faster for gossip: from 81.8\% at $N=5$ to 61.2\% at $N=40$, compared with 84.3\% to 68.2\% for FedAvg (Fig.~\ref{fig:exploratory}b). At $N=40$ each node holds only two training engines, and the ring's SLEM of 0.992 means that information from one node needs many rounds to reach distant nodes. These experiments fragment a fixed dataset and model message delivery rather than a physical network; they probe data fragmentation, not the behaviour of large deployments.

\subsection{Replication on FD002--FD004}
\label{sec:res-subsets}
\begin{figure*}[t]
\centering
\includegraphics[width=\textwidth]{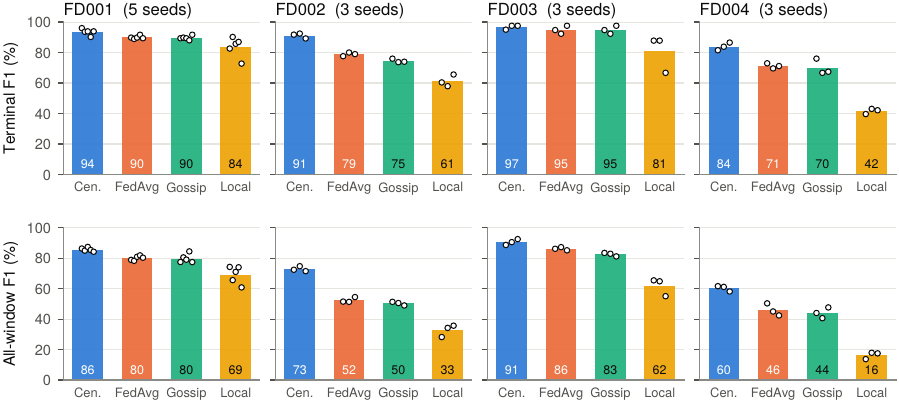}
\caption{Replication across all four C-MAPSS subsets with an unchanged protocol. Bars show the mean terminal-window (top) and all-window (bottom) F1; white dots are individual seeds.}
\label{fig:subsets}
\end{figure*}

Fig.~\ref{fig:subsets} and Table~\ref{tab:subsets} repeat the main comparison on the other three subsets without changing the protocol. The benefit of communication held on every subset: gossip exceeded local-only training on the primary endpoint by 13.3\,pp on FD002, 14.1\,pp on FD003 and 28.4\,pp on FD004, and did so in every seed. The gap between gossip and FedAvg depended on the subset. On FD003 and FD004 the two were close on terminal windows (94.9\,$\pm$\,2.6\% versus 94.9\,$\pm$\,2.6\%, and 70.0\,$\pm$\,5.2\% versus 71.2\,$\pm$\,1.7\%). On FD002, which combines six operating conditions with the largest number of engines, gossip was lower in every seed (74.5\,$\pm$\,1.2\% versus 78.8\,$\pm$\,1.2\%; $-4.3$\,pp). On all windows, gossip was also below FedAvg on FD003 in every seed (82.5\,$\pm$\,1.2\% versus 86.2\,$\pm$\,1.0\%) and on FD002 (50.3\,$\pm$\,1.2\% versus 52.5\,$\pm$\,1.8\%).

\begin{table}[t]
\centering
\caption{Results on FD002--FD004 (three seeds, mean\,$\pm$\,SD, \%). Evaluation populations are listed in Table~\ref{tab:data}.}
\label{tab:subsets}
\footnotesize
\setlength{\tabcolsep}{3pt}
\begin{adjustbox}{max width=\linewidth}
\begin{tabular}{@{}llcccc@{}}
\toprule
& & \multicolumn{2}{c}{Terminal window} & \multicolumn{2}{c}{All windows} \\
\cmidrule(lr){3-4}\cmidrule(lr){5-6}
Subset & Method & F1 & AP & F1 & AP \\
\midrule
\multirow{4}{*}{FD002} & Centralized & \pmsd{91.1}{1.7} & \pmsd{98.7}{0.4} & \pmsd{72.9}{1.7} & \pmsd{84.4}{3.2} \\
& FedAvg & \pmsd{78.8}{1.2} & \pmsd{90.3}{1.4} & \pmsd{52.5}{1.8} & \pmsd{55.2}{0.7} \\
& Ring gossip & \pmsd{74.5}{1.2} & \pmsd{85.4}{2.2} & \pmsd{50.3}{1.2} & \pmsd{50.3}{1.9} \\
& Local-only & \pmsd{61.3}{3.9} & \pmsd{66.7}{3.0} & \pmsd{32.7}{4.0} & \pmsd{24.8}{3.0} \\
\midrule
\multirow{4}{*}{FD003} & Centralized & \pmsd{96.7}{1.5} & \pmsd{99.8}{0.1} & \pmsd{90.6}{1.9} & \pmsd{98.0}{0.5} \\
& FedAvg & \pmsd{94.9}{2.6} & \pmsd{99.5}{0.3} & \pmsd{86.2}{1.0} & \pmsd{94.5}{0.4} \\
& Ring gossip & \pmsd{94.9}{2.6} & \pmsd{99.2}{0.5} & \pmsd{82.5}{1.2} & \pmsd{91.2}{0.7} \\
& Local-only & \pmsd{80.8}{12.2} & \pmsd{91.3}{1.0} & \pmsd{61.8}{5.8} & \pmsd{65.3}{10.8} \\
\midrule
\multirow{4}{*}{FD004} & Centralized & \pmsd{83.9}{2.5} & \pmsd{91.2}{2.7} & \pmsd{60.3}{1.9} & \pmsd{60.9}{4.1} \\
& FedAvg & \pmsd{71.2}{1.7} & \pmsd{79.2}{1.2} & \pmsd{45.9}{4.0} & \pmsd{48.5}{2.7} \\
& Ring gossip & \pmsd{70.0}{5.2} & \pmsd{74.7}{4.7} & \pmsd{44.1}{3.5} & \pmsd{40.4}{6.2} \\
& Local-only & \pmsd{41.6}{1.8} & \pmsd{44.5}{6.9} & \pmsd{16.4}{2.3} & \pmsd{9.3}{2.4} \\
\bottomrule
\end{tabular}
\end{adjustbox}
\end{table}

The six-condition subsets FD002 and FD004 were considerably harder for every method. The global min--max scaling used throughout does not normalize sensors per operating condition, and condition-aware preprocessing, which is common for these subsets, would be expected to raise absolute performance. We kept the FD001 protocol unchanged because the question was whether the relative ordering of the training protocols transfers, not how to maximize accuracy on each subset.

\subsection{Membership-inference diagnostic}
\label{sec:res-privacy}
On FD001, the loss-threshold membership test reached an AUC of 0.82\,$\pm$\,0.04 for local-only models, 0.76\,$\pm$\,0.06 for the centralized model, 0.60\,$\pm$\,0.07 for gossip and 0.57\,$\pm$\,0.09 for FedAvg. Membership was detectable above chance for every protocol and least so for the collaboratively trained models, whose parameters average information from many engines. The diagnostic is confounded by differences between engines and is not a privacy guarantee.

\section{Discussion}
\label{sec:discussion}

\subsection{Collaboration matters most}
The largest and most consistent effect in this study is the value of communication itself. A node that sees only eight FD001 engines learns a markedly worse and less stable detector than any collaborative protocol, and on the harder subsets the gap widens to 13--28\,pp of terminal-window F1. For a maintenance operator deciding whether to join a collaborative scheme, this is the first-order consideration: the choice between federated and serverless aggregation is secondary to the decision to collaborate at all.

\subsection{When a ring is enough}
On FD001, FD003 and FD004, ring gossip reached the same primary-endpoint accuracy as FedAvg while transmitting the same payload without a coordinator. This agrees with general empirical comparisons of gossip learning and FL \cite{hegedus2021gossip} and with decentralized SGD theory, in which a sufficiently well-mixing topology approaches centralized behaviour \cite{lian2017dpsgd,koloskova2020unified}. The limits of the ring became visible in two situations. On FD002, with six operating conditions spread randomly over nodes, gossip trailed FedAvg by 4.3\,pp in every seed, and on all windows of FD003 it trailed by 3.7\,pp. As the same data were fragmented over 20 and 40 nodes, gossip degraded faster than FedAvg. Both observations are consistent with slow mixing: in a ring of $N$ nodes the SLEM approaches one as $N$ grows, so knowledge from one node reaches distant nodes only after many rounds, and heterogeneous local updates keep pulling the models apart in the meantime.

These results suggest a practical rule for serverless PdM systems. A sparse ring is adequate when node data are moderately heterogeneous and networks are small. When operating regimes differ strongly across sites or many sites participate, the topology should mix faster, for example through additional chords, several gossip steps per round or time-varying peer selection, all of which trade extra communication for faster agreement \cite{boyd2006gossip,koloskova2020unified}. Our released code supports these variants, and quantifying this trade-off is a direct next step.

\subsection{Robustness, communication and privacy in context}
The simulated faults did not separate the protocols: both tolerated message loss and a single node failure, and FedAvg nodes simply trained locally during a server outage. The robustness benefit of gossip in these settings therefore lies in not needing a coordinator at all, which matters for organizational trust and system design, rather than in higher accuracy under faults. Total payload was identical at $N=10$, but its distribution differs: each gossip node exchanges data with two peers, whereas the FedAvg server must terminate every transfer. Communication-efficient variants deserve further study, because the pre-specified event trigger saved nothing when parameters changed by more than 2\% per epoch. Finally, keeping raw data on the nodes is not formal privacy. Shared parameters can leak information about training data \cite{zhu2019dlg,shokri2017mia}, and the membership diagnostic confirms leakage above chance for all protocols; differential privacy or secure aggregation would be required for formal guarantees.

\subsection{Limitations}
All experiments use simulated C-MAPSS data. The ten-node partitions are constructed rather than observed, and the main analyses use the single-condition FD001 subset; FD002--FD004 were evaluated with three seeds and without condition-specific preprocessing. Results may not transfer to real fleets with site-specific operating regimes, sensor faults or label noise, and evaluation on N-CMAPSS \cite{arias2021ncmapss} and industrial multi-site data is needed. The network is simulated synchronously on one CPU, so latency, asynchronous operation and energy use on edge hardware were not measured. The primary endpoint of FD001 contains only 93 test engines, which limits the resolution of comparisons between the stronger protocols. The fixed RUL horizon of 30 cycles and decision threshold of 0.5 define a single operating point, and cost-sensitive threshold selection was not studied.

\section{Conclusion}
\label{sec:conclusion}
We compared serverless ring gossip, FedAvg, isolated local training and centralized training of an LSTM imminent-failure detector on all four NASA C-MAPSS subsets under a single matched protocol with repeated seeds and an engine-level primary endpoint. Peer communication consistently and substantially improved on local training. Ring gossip matched FedAvg on FD001, FD003 and FD004 with the same payload and no coordinator, and its models agreed closely but not exactly. On the heterogeneous FD002 subset and in larger rings, slower mixing made gossip measurably less accurate than FedAvg. Within the limits of simulated data and a simulated network, ring gossip is a workable serverless option for collaborative failure detection when heterogeneity is moderate, and faster-mixing topologies should be preferred as heterogeneity and network size grow. All code, data and per-run results are released to support verification and extension.

\section*{CRediT authorship contribution statement}
\textbf{Yusuf Öztürk:} Conceptualization, Methodology, Formal analysis, Investigation, Data curation, Writing -- original draft, Supervision, Project administration.
\textbf{Enes Göktekin:} Conceptualization, Methodology, Software, Validation, Investigation, Writing -- original draft, Visualization.
\textbf{Bengisu Atlı:} Resources, Data curation, Software.
\textbf{Akın Öztürk:} Methodology, Formal analysis, Investigation, Data curation, Writing -- original draft.
\textbf{Zhixiang Wang:} Software, Validation, Formal analysis, Investigation, Visualization, Writing -- review \& editing.
\textbf{Ulas Bagci:} Conceptualization, Supervision, Project administration, Writing -- review \& editing.

\section*{Declaration of competing interest}
The authors declare that they have no known competing financial interests or personal relationships that could have appeared to influence the work reported in this paper.

\section*{Acknowledgements}
This work was supported by the National Institutes of Health (NIH) under grants R01-HL171376 and U01-CA268808. The content is solely the responsibility of the authors and does not necessarily represent the official views of the National Institutes of Health.

\section*{Declaration of generative AI and AI-assisted technologies in the manuscript preparation process}
During the preparation of this work the authors used ChatGPT (OpenAI) and Claude (Anthropic) in order to improve language clarity, restructure and edit the manuscript, check reference metadata, and assist with code review and analysis scripts. After using these tools, the authors reviewed and edited the content as needed and take full responsibility for the content of the published article.

\section*{Data availability}
The NASA C-MAPSS dataset is publicly available \cite{saxena2008cmapss}. The code, the C-MAPSS data files used, experiment configurations, per-run metrics and training histories, and the scripts that regenerate all predictions, tables and figures are available at \url{https://github.com/ZhixiangWang-CN/gossip-lstm-cmapss}.

\bibliographystyle{elsarticle-num}
\bibliography{refs}

\end{document}